\documentclass[10pt,twocolumn,letterpaper]{article}
\usepackage{3dv}
\usepackage{algorithm}
\usepackage{algorithmic}
\usepackage{amsmath}
\usepackage{amssymb}
\usepackage{array}
\usepackage{balance}
\usepackage{bm}
\usepackage{booktabs}
\usepackage{cite}
\usepackage{color}
\usepackage{colortbl} 
\usepackage{cuted}
\usepackage{epsfig}
\usepackage{float}
\usepackage[T1]{fontenc}
\usepackage{graphicx}
\usepackage[colorlinks = true,
linkcolor = black,
urlcolor  = magenta,
citecolor = black,
anchorcolor = black,
pagebackref=true,
breaklinks=true,
bookmarks=false
]{hyperref}
\usepackage{makecell}
\usepackage{mathrsfs}
\usepackage{mathtools}
\usepackage{multirow}
\usepackage{stfloats}
\usepackage{tabulary}
\usepackage{times}
\usepackage{url}
\usepackage[table]{xcolor}
\threedvfinalcopy
\ifthreedvfinal\fi

\begin{document}
	\title{SDA-SNE: Spatial Discontinuity-Aware Surface Normal Estimation via Multi-Directional Dynamic Programming}
	\author{Nan Ming, Yi Feng, Rui Fan\thanks{Corresponding author}\\
		MIAS Research Group, Robotics \& Artificial Intelligence Laboratory,\\
		Department of Control Science \& Engineering,\\Tongji University, Shanghai 201804, China\\
		{\tt\small \{nan.ming, fengyi, rui.fan\}@ieee.org}
	}
	
	\maketitle
	\thispagestyle{empty}

	\begin{abstract}
		The state-of-the-art (SoTA) surface normal estimators (SNEs) generally translate depth images into surface normal maps in an end-to-end fashion. Although such SNEs have greatly minimized the trade-off between efficiency and accuracy, their performance on spatial discontinuities, \textit{e.g.}, edges and ridges, is still unsatisfactory. To address this issue, this paper first introduces a novel multi-directional dynamic programming strategy to adaptively determine inliers (co-planar 3D points) by minimizing a (path) smoothness energy. The depth gradients can then be refined iteratively using a novel recursive polynomial interpolation algorithm, which helps yield more reasonable surface normals. Our introduced spatial discontinuity-aware (SDA) depth gradient refinement strategy is compatible with any depth-to-normal SNEs. Our proposed SDA-SNE achieves much greater performance than all other SoTA approaches, especially near/on spatial discontinuities. We further evaluate the performance of SDA-SNE with respect to different iterations, and the results suggest that it converges fast after only a few iterations. This ensures its high efficiency in various robotics and computer vision applications requiring real-time performance. Additional experiments on the datasets with different extents of random noise further validate our SDA-SNE’s robustness and environmental adaptability.  Our source code, demo video, and supplementary material are publicly available at \url{mias.group/SDA-SNE}.
	\end{abstract}
	
	\section{Introduction}
	\label{sec:intro}
	Surface normal is an informative 3D visual feature used in various computer vision and robotics applications, such as object recognition and scene understanding \cite{PAMI2012, klasing2009comparison,fan2019pothole}. To date, there has not been extensive research on surface normal estimation, as it is merely considered to be an auxiliary functionality for other computer vision and robotics applications \cite{3F2N}. As such applications are typically required to perform robustly and in real time, surface normal estimators (SNEs) must be sufficiently accurate and computationally efficient \cite{3F2N}. 
	
	\begin{figure}[!t] 
		\centering
		\includegraphics[width=0.48\textwidth]{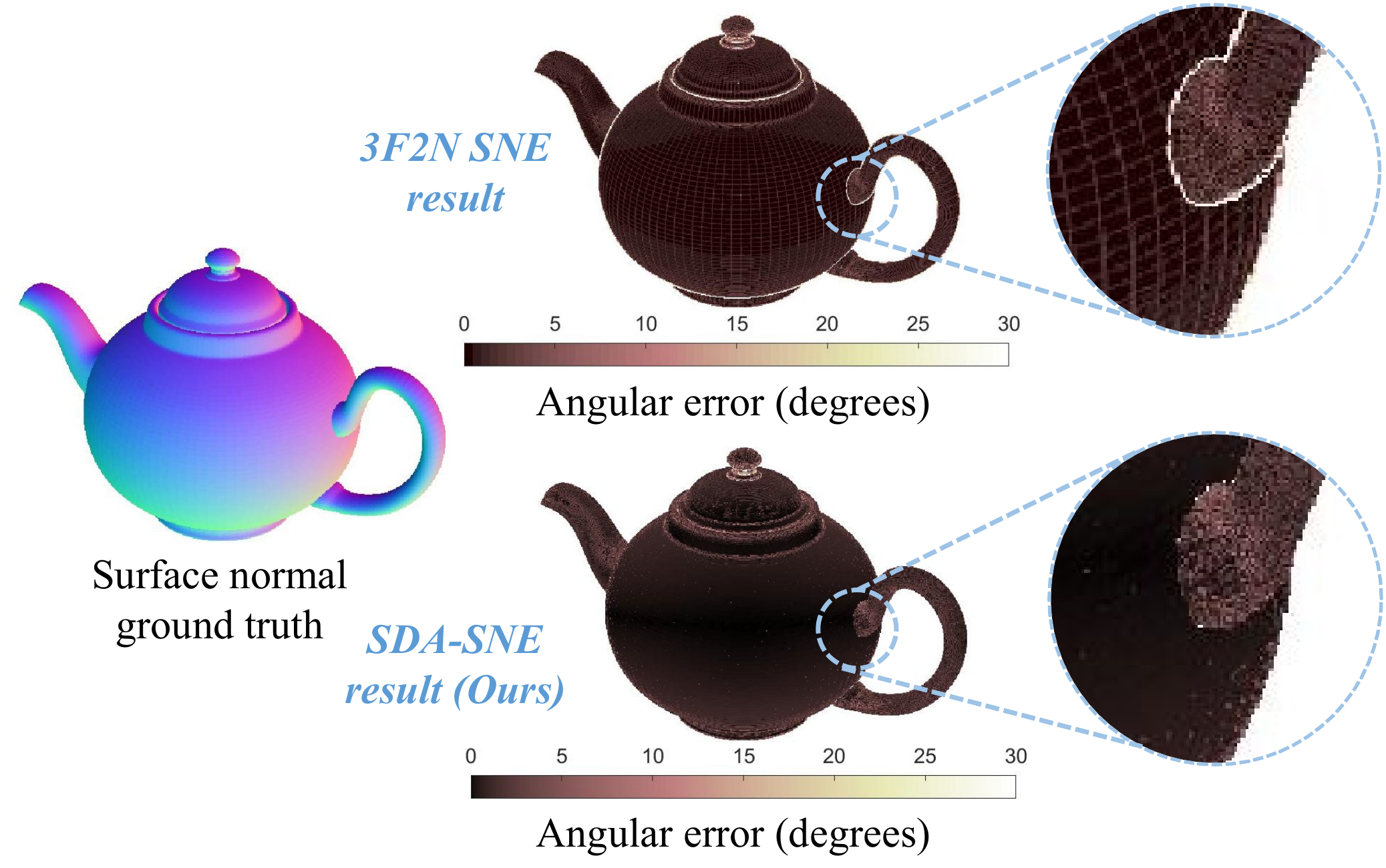}
		\vspace{+0.5em}
		\caption{Comparison between 3F2N (recently published SoTA SNE) \cite{3F2N} and our proposed SDA-SNE. A significantly improved region is marked with a dashed blue circle.}
		\label{fig.front}
	\end{figure}
	
	The state-of-the-art (SoTA) SNEs \cite{klasing2009realtime,jordan2014quantitative,klasing2009comparison,3F2N,CP2TV,raposo2020accurate,daniel2015optimal,barath2019optimal} generally select a set of 3D points and compute surface normals via planar fitting, geometric transformation, or statistical analysis. However, such approaches are infeasible to estimate surface normals near/on spatial discontinuities, \eg, edges and ridges (see Fig. \ref{fig.front}), as they generally introduce adjacent 3D points on different surfaces unintentionally \cite{JV2018}. Bormann \emph{et al.} \cite{IROS2015} proposed the first edge-aware SNE, capable of adaptively selecting reasonable adjacent 3D points on the same surface. It performs significantly better than prior arts near edges. Nonetheless, their method focuses only on edges and requires a manually-set discontinuity awareness threshold. This results in low adaptability to different scenarios and datasets. Hence, there is a strong motivation to develop an SNE capable of tackling all types of spatial discontinuities to achieve greater robustness and environmental adaptability.
	
	The major contributions of this work are summarized as follows:\\
	(1) \textbf{Spatial discontinuity-aware surface normal estimator (SDA-SNE)}, a highly accurate SNE, significantly outperforming all other SoTA SNEs, especially near/on spatial discontinuities. \\
	(2) \textbf{Multi-directional dynamic programming (DP)} for depth gradient refinement. It iteratively introduces inliers (co-planar pixels) by minimizing a smoothness energy.\\
	(3) \textbf{Path discontinuity (PD) norm}, a semi-norm evaluating the discontinuity of a given path. It reflects the depth gradient estimation error generated by the Taylor expansion.\\
	(4) \textbf{Recursive polynomial interpolation (RPI)}, an ultrafast polynomial interpolation algorithm. Compared to Lagrange and Newton polynomial interpolation, it iteratively refines the first-order derivative of depth with a more efficient polynomial interpolation strategy.
	\section{Related Work}
	\label{sec:RelatedWork}
	
	The existing SoTA SNEs can be categorized into four classes: (1) optimization-based \cite{klasing2009comparison}, (2) averaging-based \cite{klasing2009realtime}, (3) affine-correspondence-based \cite{barath2019optimal}, and (4) depth-to-normal \cite{wang2021sne}.
	
	The optimization-based SNEs, \eg, PlaneSVD \cite{klasing2009realtime}, PlanePCA \cite{jordan2014quantitative}, VectorSVD \cite{klasing2009realtime}, and QuadSVD \cite{2001Cloud,2005On}, compute surface normals by fitting local planar or curved surfaces to an observed 3D point cloud, using either singular value decomposition (SVD) or principal component analysis (PCA). The averaging-based SNEs, \eg, AreaWeighted \cite{klasing2009comparison} and AngleWeighted \cite{klasing2009realtime}, estimate surface normals by computing the weighted average of the normal vectors of the triangles formed by each given 3D point and its neighbors. However, these two categories of SNEs are highly computationally intensive and unsuitable for online robotics and computer vision applications \cite{3F2N}. The affine-correspondence-based SNEs exploit the relationship between affinities and surface normals \cite{daniel2015optimal, raposo2020accurate, barath2019optimal}. Nevertheless, such SNEs are developed specifically for stereo or multi-view cases and cannot generalize to other cases where monocular depth images are used.

	Recently, enormous progress has been made in end-to-end depth-to-normal translation \cite{3F2N, fan2020sne, CP2TV}. Such SNEs have demonstrated superior performance in terms of both speed and accuracy. Fan \emph{et al.} \cite{3F2N} proposed a fast and accurate SNE referred to as 3F2N, which can infer surface normal information directly from depth or disparity images, with two gradient filters and one central tendency measurement filter (a mean or median filtering operation), as follows:
	\begin{equation}
		\begin{split}
			&{n_x= {\partial d}/{\partial u}, \ \ \ \ \ \ \ \ \ \ \  n_y= {\partial d}/{\partial v}},\\
			&{{n}_z}=-\Phi\Bigg\{ \frac{({x_i}-x) n_x + ({y_i}-y) n_y }{{z_i}-z} \Bigg\},\ i = 1,\dots,m,
			\label{eq.3F2N}
		\end{split}
	\end{equation}
	where $\mathbf {n} = [n_x,n_y,n_z]^\top$ is the surface normal of a given 3D point $\mathbf{p}^C= [x,y,z]^\top$ in the camera coordinate system, $\mathbf{p}^C$ is projected to a 2D pixel $ \mathbf {p}  = [u, v]^\top $ via $z[{\mathbf {p}}^\top,1]^\top=\mathbf{K}\mathbf{p}^C$ ($\mathbf{K}$ is the camera intrinsic matrix), $\mathbf{p}_i^C=[x_i,y_i,z_i]^\top$ is one of the $m$ adjacent pixels of $\mathbf{p}^C$, $d$ is disparity (or inverse depth), and  $\Phi\{\cdot\}$ represents the central tendency measurement filtering operation for $n_z$ estimation. Nakagawa \emph{et al.} \cite{CP2TV} presented an SNE based on cross-product of two tangent vectors (CP2TV): $\mathbf n(u,v)=\bm t_u\times \bm t_v$, where
	\begin{equation}
		\begin{split}
			\bm t_u & = \begin{bmatrix}\begin{displaystyle}\frac{\partial x}{\partial u},\frac{\partial y}{\partial u},\frac{\partial z}{\partial u}\end{displaystyle}\end{bmatrix}^\top \\ 
			\bm t_v & = \begin{bmatrix}\begin{displaystyle}\frac{\partial x}{\partial v},\frac{\partial y}{\partial v},\frac{\partial z}{\partial v}\end{displaystyle}\end{bmatrix}^\top,
		\end{split}
		\label{eq.CP2TV.1}
	\end{equation}
	\begin{equation}
		\begin{split}
			\frac{\partial x}{\partial u} & = \frac{z}{f_u}+\frac{u-c_u}{f_u}\frac{\partial z}{\partial u}\\
			\frac{\partial y}{\partial u} & = \frac{v-c_v}{f_v}\frac{\partial z}{\partial u}\\
			\frac{\partial x}{\partial v} & = \frac{u-c_u}{f_u}\frac{\partial z}{\partial v}\\
			\frac{\partial y}{\partial v} & = \frac{z}{f_v}+\frac{v-c_v}{f_v}\frac{\partial z}{\partial v},
		\end{split}
		\label{eq.CP2TV.2}
	\end{equation}
	$\mathbf{c}=[c_u,c_v]^\top$ is the principal point (in pixels), and $f_u$ and $f_v$ are the camera focal lengths (in pixels) in the $u$ and $v$ directions, respectively. Since the depth or disparity gradient estimation has a direct impact on the surface normal quality, this paper focuses thoroughly on the strategy to improve the accuracy of depth gradient $\nabla \hat z = [ \hat z_{u},\hat z_{v} ]^{\top}$\footnote{In this paper, the variables with and without hat symbols denote the estimated and theoretical values, respectively.}. The proposed strategy can also be applied to improve disparity gradient estimation, as it is inversely proportional to depth.

	\section{Methodology}
	\label{sec:Methodology}
	As discussed in Sec. \ref{sec:RelatedWork}, surface normal estimation was formulated as a depth-to-normal translation problem in SoTA approaches, which have demonstrated superior performances over other methods in terms of both speed and accuracy \cite{3F2N}. The accuracy of such approaches is subject to the depth gradient quality. Since depth gradients typically jump near/on discontinuities, \eg, edges and ridges, the estimated surface normals near/on such discontinuities are always significantly different from their ground truth. Therefore, we propose a multi-directional DP strategy to translate a depth or disparity image into a surface normal map in a coarse-to-fine manner. It first initializes $\nabla \hat z$ using finite difference interpolation and then optimizes $\nabla \hat z$ using our proposed multi-directional DP. Accurate surface normals can then be obtained by plugging the optimum depth gradients into either the central tendency measurement filtering operation used in 3F2N (see (\ref{eq.3F2N})) or the cross-product operation used in CP2TV (see (\ref{eq.CP2TV.1})-(\ref{eq.CP2TV.2})).

	\begin{figure*}[!t]  
		\centering
		\includegraphics[width=1\textwidth]{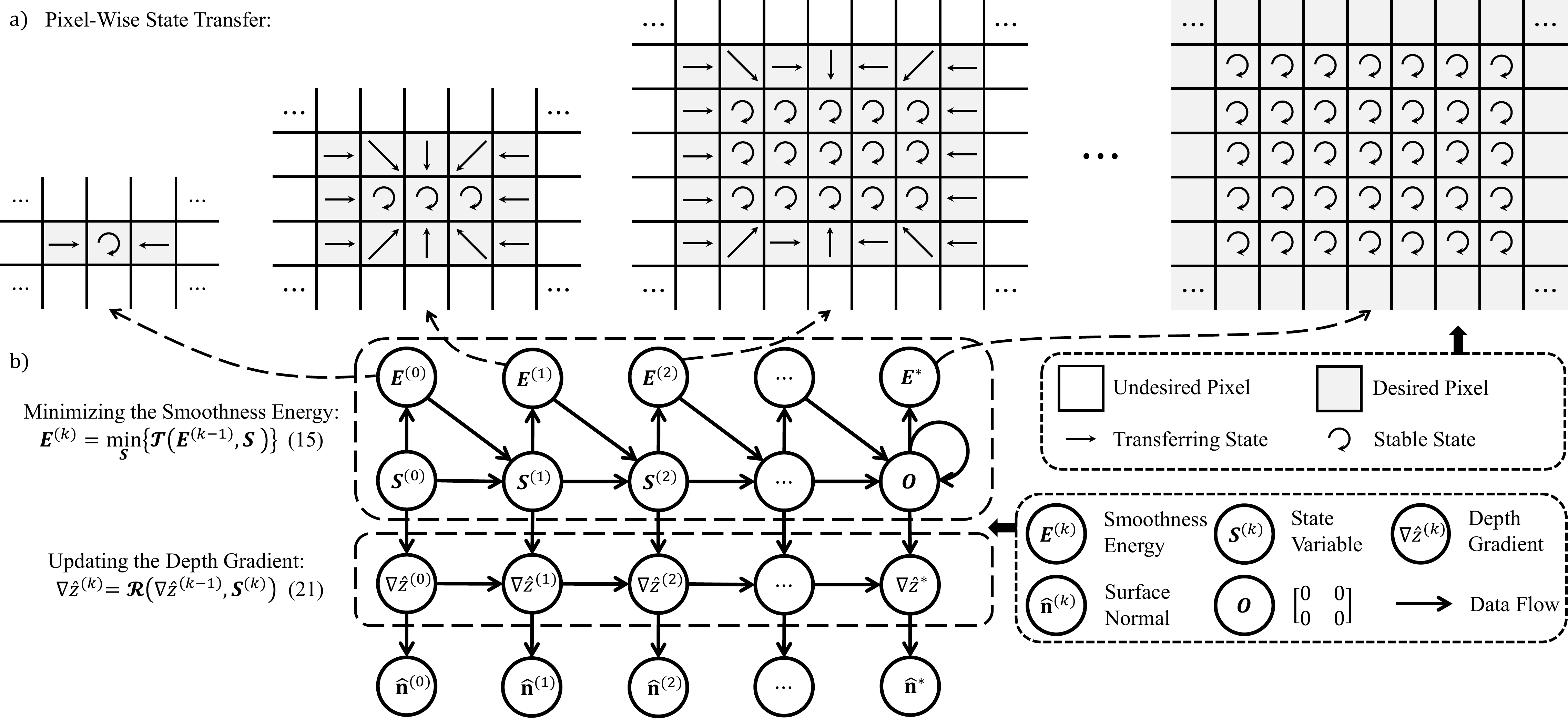}
		\caption{
			(a) An illustration of the energy transfer process, where the inliers (desired pixels) are gradually introduced in each iteration; (b) The relationship among smoothness energy, state variable, depth gradient, and surface normal.}
		\label{fig.dp}
	\end{figure*}

	\subsection{Multi-Directional Dynamic Programming}
	\subsubsection{Smoothness Energy and State Variable}
	As a general rule, only inliers (co-planar pixels) should be introduced when estimating the depth gradient of an observed pixel ${\mathbf {p}}$. Such pixels are determined by minimizing the (path) smoothness energy $\bm E=[E_u,E_v]$ of ${\mathbf {p}}$ using DP, where $E_u$ and $E_v$ represent the horizontal and vertical components of $\bm E$, respectively. To minimize the computational complexity, we recursively introduce adjacent co-planar pixels. In each iteration, two desired pixels ${\mathbf {p}}_u'$ and ${\mathbf {p}}_v'$ (corresponding to the minimum $E_u$ and $E_v$, respectively) are used to extend the DP path and  update the depth gradient of ${\mathbf {p}}$. We define a state variable $\bm S=[\bm s_u,\bm s_v]$, where $\bm s_u={\mathbf {p}}_u'-{\mathbf {p}}$ and $\bm s_v={\mathbf {p}}_v'-{\mathbf {p}}$ denote the horizontal and vertical components of $\bm S$, respectively\footnote{$\bm s_u,\bm s_v \in \big\{[i,j]^{\top}\mid i,j\in \{-1,0,1\}\big\}$ because only adjacent pixels are considered.}. In the $k$-th iteration, the desired pixels are determined by minimizing $\bm E^{\left(k\right)}$ as follows:
	\begin{equation}
		\bm E^{\left(k\right)}=\min\limits_{\bm S}\left\{\bm{\mathcal T}\left(\bm E^{\left(k-1\right)},\bm S \right)\right\}.
		\label{def.C1}
	\end{equation}
	The state variable $\bm S^{\left(k\right)}$ is transferred to
	\begin{equation} \bm S^{\left(k\right)}=\mathop{\arg\min}\limits_{\bm S}\left\{\bm{\mathcal T}\left(\bm E^{\left(k-1\right)},\bm S \right)\right\},
		\label{def.C2}
	\end{equation}
	where $\bm {\mathcal T}$ denotes the smoothness energy transfer function. Therefore, we can update the depth gradient with the desired pixels using:
	\begin{equation}
		\nabla \hat z^{\left(k\right)}=\bm{\mathcal R}\left(\nabla \hat z^{\left(k-1\right)},\bm S^{\left(k\right)}\right),
		\label{def.R}
	\end{equation}
	where $\bm {\mathcal R}$ denotes the depth gradient update function. 

As shown in Fig. \ref{fig.dp}, after the initialization of depth gradient $\nabla \hat{z}$ and DP variables $\bm E$ and $\bm S$ (detailed in Sec. \ref{sec:Ini}), inliers are iteratively introduced using the smoothness energy transfer function $\bm{\mathcal T}$ (detailed in Sec. \ref{sec:C}) and the depth gradients are iteratively updated using the depth gradient update function $\bm{\mathcal R}$ (detailed in Sec. \ref{sec:R}). The optimum depth gradients can be obtained when the smoothness energy converges to the minimum. The pseudo-code of multi-directional DP is given in Algorithm \ref{alg:DP}.

		\begin{figure}[!t]
		
		\renewcommand{\algorithmicrequire}{\textbf{Input:}}
		\renewcommand{\algorithmicensure}{\textbf{Output:}}
		\begin{algorithm}[H]
			\caption{Multi-Directional DP}
			\label{alg:DP}
			\begin{algorithmic}[1]
				\REQUIRE Depth map $z$, Initial depth gradient $[\hat z_{u}^{\left(0\right)},\hat z_{v}^{\left(0\right)}]$, Initial smoothness energy components $E^{\left(0\right)}_u$ and $E^{\left(0\right)}_v$, Initial state variable components $\bm s_u^{\left(0\right)}$ and $\bm s_v^{\left(0\right)}$.    
				\ENSURE Optimum depth gradient $[\hat z_{u}, \hat z_{v}]$
				\STATE {$[\hat z_{u},\hat z_{v}] \leftarrow [\hat z_{u}^{\left(0\right)},\hat z_{v}^{\left(0\right)}]$, \\
					$[E_{u},E_{v}] \leftarrow [E^{\left(0\right)}_u, E^{\left(0\right)}_v]$,\\
					$[\bm s_u,\bm s_v] \leftarrow [\bm s_u^{\left(0\right)},\bm s_v^{\left(0\right)}]$}
				\REPEAT
				\STATE {Initialize energy set $\Omega_u \leftarrow \varnothing,\Omega_v \leftarrow \varnothing$}
				\FOR {\textbf{each} ${\mathbf {p}}'$}
				\STATE {$\Omega_u[{\mathbf {p}}] \!\leftarrow \! \Omega_u[{\mathbf {p}}] \cup \mathcal T(E_u,E_v,{\mathbf {p}},{\mathbf {p}}',\bm s_u$})
				\STATE {$\Omega_v[{\mathbf {p}}] \leftarrow \Omega_v[{\mathbf {p}}] \cup \mathcal T(E_u,E_v,{\mathbf {p}},{\mathbf {p}}',\bm s_v$})\\
				\ENDFOR
				\STATE {$\bm s_u[{\mathbf {p}}] \leftarrow \mathop{\arg\min}\limits_{{\mathbf {p}}'\in \mathcal N_1({\mathbf {p}})}\left\{\Omega_u[{\mathbf {p}}][{\mathbf {p}}']\right\} - {\mathbf {p}}$}
				\STATE {$\bm s_v[{\mathbf {p}}] \leftarrow \mathop{\arg\min}\limits_{{\mathbf {p}}'\in \mathcal N_1({\mathbf {p}})}\left\{\Omega_v[{\mathbf {p}}][{\mathbf {p}}']\right\} - {\mathbf {p}}$}
				\STATE {$E_u[{\mathbf {p}}] \leftarrow \Omega_u[{\mathbf {p}}][{\mathbf {p}}+\bm s_u] $}
				\STATE {$E_v[{\mathbf {p}}] \leftarrow \Omega_v[{\mathbf {p}}][{\mathbf {p}}+\bm s_v] $}
				\STATE {$[\hat z_{u},\hat z_{v}] \leftarrow [\mathcal R(\hat z_{u},\hat z_{v}, \bm s_u),\mathcal R(\hat z_{u},\hat z_{v}, \bm s_v)]$}
				\UNTIL{all ($\bm s_u[:]=\bm 0$) \textbf{and} all ($\bm s_v[:]=\bm 0$)}
				\RETURN{$[\hat z_{u}, \hat z_{v}]$}
			\end{algorithmic}
		\end{algorithm}
 	\end{figure}
	
\subsubsection{Initialization of Depth Gradient, Smoothness Energy, and State Variable}
	\label{sec:Ini}
	In the initialization stage, the coarse depth gradient $\nabla \hat z^{\left( 0 \right)}$ of a given pixel is obtained by computing the finite difference (FD) between adjacent pixels. The FDs of depth are divided into $\Delta_f z = [\Delta_f z_u,\Delta_f z_v]^{\top}$ and $\Delta_b z = [\Delta_b z_u, \Delta_b z_v]^{\top}$, where the subscripts $u$ and $v$ represent the horizontal and vertical directions, respectively, and $\Delta_f$ and $\Delta_b$ represent forward and backward FDs, respectively. In order to achieve the discontinuity awareness ability, we set different weights to these FD operators by comparing the local smoothness measured by $\hat z_{uu}$ and $\hat z_{vv}$, the second-order partial derivatives of adjacent pixels, as follows:
	\begin{equation}
		\begin{split}
			\bm \eta=\begin{bmatrix}\eta_u,\eta_v\end{bmatrix}=\mathop{\arg\min}\limits_{[i,j]}\left\{|\hat z_{uu}(u+i,v)|+|\hat z_{vv}(u,v+j)|\right\}\\
		\end{split},
		\label{eq.ini.xy}
	\end{equation}
	where $i,j\in \{-1,0,1\}$. We can then use the smoothest pixels to linearly interpolate $\Delta_f z$ and $\Delta_b z$:
	\begin{equation}
		\begin{split}
			\nabla \hat z^{\left( 0 \right)}=\frac{1}{2}(\bm 1+\bm \eta)^{\top} \circ \Delta_f z + \frac{1}{2}(\bm 1-\bm \eta)^{\top} \circ \Delta_b z
		\end{split},
		\label{eq.ini.G}
	\end{equation}
	where $\circ$ denotes the Hadamard product and $\bm 1=[1,1]^{\top}$. 
	Furthermore, we initialize the smoothness energy $\bm E^{\left(0\right)}$ as $\begin{bmatrix}\min\{|\hat z_{uu}(u+i,v)|\}, \min\{ |\hat z_{vv}(u,v+j)|\}\end{bmatrix}$ and the state variable $\bm S^{\left(0\right)}$ as $\mathrm{diag}(\eta_u,\eta_v)$.
	
	\subsection{Smoothness Energy Transfer Function}
	\label{sec:C}
This section discusses the formulation of the smoothness energy transfer function $\bm{\mathcal T}$. Energy transfer can be divided into two categories: 1) collinear transfer ($\bm s_u \times \mathbf e_1 = \bm 0$ or $\bm s_v \times \mathbf e_2 = \bm 0$, where $\mathbf e_1=[1,0]^\top$ and $\mathbf e_2=[0,1]^\top$ are the unit orthogonal base of 2D Euclidean space); 2) non-collinear transfer ($\bm s_u \times \mathbf e_1 \neq \bm 0$ or $\bm s_v \times \mathbf e_2 \neq \bm 0$). As a general rule, the collinear pixels should be introduced to adapt to the convex surface if they are smooth enough; otherwise, non-collinear pixels should be introduced to deal with discontinuities. 
	
	\subsubsection{Depth Gradient Calculation}
	Based on the gradient theorem \cite{williamson2004multivariable}, the relationship among $z$, $z_u$, and $z_v$ can be written as follows:
	\begin{equation}
		z({\mathbf {p}}')-z(\mathbf {p})=\int_{\mathcal L}z_u\,du+z_v\,dv,
		\label{eq.PathIntegral.0}
	\end{equation}
	where $\mathbf {p} = [u_0,v_0]^{\top}$ is the given pixel, ${\mathbf {p}}' = [u_1,v_1]^{\top}$ is its adjacent pixel to be introduced (satisfying $|u_0-u_1| \le 1, |v_0-v_1| \le 1$), and $\mathcal L$ is the path from $\mathbf {p}$ to ${\mathbf {p}}'$. We can therefore obtain (i) the collinear transfer as follows:
	\begin{equation}
		\begin{split}
			& z({\mathbf {p}}')-z(\mathbf {p})=\int_{u_0}^{u_1} z_u\,du, 	\mbox{ or } \\
			& z({\mathbf {p}}')-z(\mathbf {p})=\int_{v_0}^{v_1} z_v\,dv,
		\end{split}
		\label{eq.PathIntegral.1}
	\end{equation}
	and (ii) the non-collinear transfer as follows:
	\begin{equation}
		z({\mathbf {p}}')-z(\mathbf {p})=\int_{\mathcal L_1}z_u\,du+z_v\,dv=\int_{\mathcal L_2}z_u\,du+z_v\,dv,
		\label{eq.PathIntegral.2}
	\end{equation}
	where $\mathcal L_1$ and $\mathcal L_2$ are two different paths from $\mathbf {p}$ to ${\mathbf {p}}'$. 
	
	In practice, however, the discrete $\hat z_{u}$ and $\hat z_{v}$ can result in errors in (\ref{eq.PathIntegral.1}) and (\ref{eq.PathIntegral.2}). The errors of $\hat z_{u}$ and $\hat z_{v}$ can be estimated using the path integral of $z_{uu}$ and $z_{vv}$ based on the Taylor expansion (more details are provided in the supplement). Hence, in order to measure these errors in a more convenient way, we propose the PD norm, which denotes the path integral (sum) of a series of $|z_{uu}|$ and $|z_{vv}|$. 
	
	\subsubsection{Path Discontinuity Norm}
	Let $z(u,v)$ be a function in $\mathbb{R}^2$ defined on an open set $\Omega$ and $\mathcal L$ be a specific path contained in $\Omega$. The discontinuity extent of the path can be measured using our proposed PD norm as follows:
	\begin{equation}
		||z||_{\text{PD}}= \int_{\mathcal L} |z_{uu}\,du|+|z_{vv}\,dv|.
		\label{def.PDN}
	\end{equation}
	After computing the PD norm, we can judge whether an adjacent pixel ${\mathbf {p}}'$ should be introduced. Therefore, we define the smoothness energy component as the PD norm with respect to a given path. In each iteration, we introduce two adjacent pixels which respectively minimize the smoothness energy components as follows:
	\begin{equation}
		\vspace{-3pt}
		\begin{split}
			& \bm s_u,\bm s_v=\mathop{\arg\min}\limits_{{\mathbf {p}}'\in \mathcal N_1({\mathbf {p}})}\left\{||z||_{\text{PD}}\right\}-\mathbf {p},\\
			\mathcal L : 
			& \begin{cases}
				\mathbf {p} \to {\mathbf {p}}' 
				&,\mbox{if }\bm s_u \times \mathbf e_1 = \bm 0\,(\mbox{or }\bm s_v \times \mathbf e_2 = \bm 0) \\
				\mathbf {p} \to {\mathbf {p}}' \to \mathbf {p} &,\mbox{if }\bm s_u \times \mathbf e_1 \neq \bm 0 \,(\mbox{or }\bm s_v \times \mathbf e_2 \neq \bm 0),
			\end{cases}
		\end{split}
		\label{eq.PDN.xy}
	\end{equation}
	where $\mathcal N_1({\mathbf {p}})=\{{\mathbf {p}}':||{\mathbf {p}}'-\mathbf {p}||_\infty \le 1\}$, and the smoothness energy components are minimized accordingly:
	\begin{equation}
		\begin{split}
			& E_u({\mathbf {p}}),E_v({\mathbf {p}})=\min\limits_{{\mathbf {p}}'\in \mathcal N_1({\mathbf {p}})}\left\{||z||_{\text{PD}}\right\}\\
			\mathcal L : 
			& \begin{cases}
				\mathbf {p} \to {\mathbf {p}}' 
				&,\mbox{if }\bm s_u \times \mathbf e_1 = \bm 0\,(\mbox{or }\bm s_v \times \mathbf e_2 = \bm 0) \\
				\mathbf {p} \to {\mathbf {p}}' \to \mathbf {p} &,\mbox{if }\bm s_u \times \mathbf e_1 \neq \bm 0 \,(\mbox{or }\bm s_v \times \mathbf e_2 \neq \bm 0).
			\end{cases}
		\end{split}
		\label{eq.PDN.C}
	\end{equation}
	The determination of co-planar pixels is, therefore, converted into an energy minimization problem.
	
	\subsubsection{Energy Minimization Strategy}
	Since depth is discrete, we formulate the energy transfer function $\bm{\mathcal T}$ in (\ref{def.C1}) as follows:
	\begin{equation}
		\begin{split}
			& E_u^{\left( k \right)}({\mathbf {p}}),E_v^{\left( k \right)}({\mathbf {p}})=\min\limits_{{\mathbf {p}}'\in \mathcal N_1({\mathbf {p}})}\Big\{E^{\left( k-1 \right)}_p({\mathbf {p}}),\\
			& |\hat z_{pp}({\mathbf {p}}'_p)|+\mathbb{I}({\mathbf {p}},{\mathbf {p}}'_p), \\
			& 2 \cdot\left[|\hat z_{oo}({\mathbf {p}}'_o)|+E^{\left( k-1 \right)}_p({\mathbf {p}}'_o)
			\right],\\
			& |\hat z_{oo}({\mathbf {p}}'_o)|+E^{\left( k-1 \right)}_p({\mathbf {p}}'_o)+|\hat z_{pp}({\mathbf {p}}'_d)|+E^{\left( k-1 \right)}_o({\mathbf {p}}'_d)  
			\Big\},
		\end{split}
		\label{eq.C}
	\end{equation}
	where the subscripts $p$, $o$, and $d$ respectively denote the variable which is parallel, orthogonal, and diagonal to the given axis (either $u$-axis or $v$-axis); the representations of $E_p$, $E_o$, $\hat z_{pp}$, $\hat z_{oo}$, $\mathbf {p}'_p$, $\mathbf {p}'_o$, and $\mathbf {p}'_d$ are given in the supplement; and the indicator function
		\begin{equation}
		\mathbb{I}({\mathbf {p}},{\mathbf {p}}'_p)=\\
		\begin{cases}
			0, 
			& \mbox{if } (\hat z_{pp}({\mathbf {p}}+\bm s_p)\cdot \hat z_{pp}({\mathbf {p}}'_p+\bm s_p) > 0 ) \\
			&\wedge (\bm s_p({\mathbf {p}})=\bm s_p^{\left( k-1 \right)}({\mathbf {p}}'_p))\\
				\infty, 
			& \mbox{otherwise } 
		\end{cases}
		\label{eq.indicator}
	\end{equation}
	add constraints to the monotonicity and convexity of the smoothness energy transfer function, improving the DP adaptivity to convex surfaces. The four terms in (\ref{eq.C}), in turn, denote the smoothness energy components when we introduce 1) no other pixels, 2) a parallel pixel ${\mathbf {p}}'_p$, 3) an orthogonal pixel ${\mathbf {p}}'_o$, and 4) a diagonal pixel ${\mathbf {p}}'_d$.

	\subsection{Depth Gradient Update Function}
	\label{sec:R}
	This subsection discusses the formulation of the depth gradient update function $\bm{\mathcal R}$. Similar to $\bm{\mathcal T}$, depth gradient update can be divided into two categories: 1) collinear update ($\bm s_u \times \mathbf e_1 = \bm 0$ or $\bm s_v \times \mathbf e_2 = \bm 0$) and 2) non-collinear update ($\bm s_u \times \mathbf e_1 \neq \bm 0$ or $\bm s_v \times \mathbf e_2 \neq \bm 0$). For the collinear update, we use all the collinear pixels which satisfy the constraint in (\ref{eq.indicator}) to estimate depth gradients, while on the other hand, for the non-collinear update, we replace the depth gradient of each given pixel with the ones on other smoother paths using (\ref{eq.PathIntegral.2}).

	\subsubsection{Depth Gradient Collinear Update with Recursive Polynomial Interpolation}
	
	The optimum $\nabla \hat z$ can be obtained by interpolating the depth of $n$ collinear pixels into an $(n-1)$-th order polynomial and subsequently computing its derivative. Lagrange or Newton polynomial interpolation has redundant computations, as each pixel is repeatedly used for polynomial interpolation. Furthermore, it is incredibly complex to yield the closed-form solution of the polynomial's derivative. To simplify the depth gradient collinear update process, we introduce a novel RPI algorithm, which can update $\nabla \hat z$ recursively with only adjacent pixels until the farthest pixel is accessed.
	
	As $\hat z_u$ and $\hat z_v$ can be computed in the same way, we only provide the details on $\hat z_u$ computation in this paper. To compute $\hat z_u$ of a given pixel $\mathbf {p} = [u_0,v_0]^{\top}$ in the $k$-th iteration, we use a $(k+1)$-th order polynomial to interpolate the depth of $(k+2)$ pixels, which have the same vertical coordinates $v_0$. The $(k+1)$-th order polynomial $f_0^{\left( k+1 \right)}(u)$ interpolated by $\{(u_0,z(u_0)),(u_0+1,z(u_0+1)),\dots,(u_0+k+1,z(u_0+k+1))\}$ can be expanded into a recursive form as follows:
	\begin{equation}
		f_0^{\left( k+1 \right)}(u)=\frac{u_0+k+1-u}{k+1}f_0^{\left( k \right)}+\frac{u-u_0}{k+1}f_1^{\left( k \right)}, 
		\label{eq.interp.1}
	\end{equation}
	where the $k$-th order polynomials $f_0^{\left( k \right)}$ and $f_1^{\left( k \right)}$ are respectively interpolated by the first and the last $k+1$ pixels (the proof of the theorem is provided in the supplement). 
	
	Since $\hat z_{u}^{\left(k-1\right)}(u_0)=\frac{df_0^{\left( k \right)}}{du}(u_0)$ and $\hat z_{u}^{\left(k-1\right)}(u_0+1)=\frac{df_1^{\left( k \right)}}{du}(u_0+1)$ have been computed in the last iteration,  we can update $\hat z_{u}(u_0)$ using:
	\begin{equation}
		\begin{split}
			& \hat z_{u}^{\left(k\right)}(u_0)=\frac{df_0^{\left( k+1 \right)}}{du}(u_0)\\
			& =\hat z_{u}^{\left(k-1\right)}(u_0)-\frac{1}{k+1}\cdot\left[z(u_0)-f_1^{\left( k \right)}(u_0)\right]. 
		\end{split}
		\label{eq.interp.2}
	\end{equation}
	Substituting $f_1^{\left( k \right)}(u_0)=z(u_0+1)-\frac{df_1^{\left( k \right)}}{du}(u_0+1)=z(u_0+1)-\hat z_{u}^{\left(k-1\right)}(u_0+1)$ into (\ref{eq.interp.2}) yields
	\begin{equation}
		\begin{split}
			& \hat z_{u}^{\left(k\right)}(u_0)=\hat z_{u}^{\left(k-1\right)}(u_0)-\\
			& \frac{1}{k+1}\cdot\left[z(u_0)-z(u_0+1)+\hat z_{u}^{\left(k-1\right)}(u_0+1)\right].
		\end{split}
		\label{eq.interp.3}
	\end{equation}
	
	\subsubsection{Depth Gradient Non-Collinear Update}
	Replacing the integral in  (\ref{eq.PathIntegral.2}) with summation yields:
	\vspace{5pt}
	\begin{equation}
		\hat z_p({\mathbf {p}}) \pm \hat z_o({\mathbf {p}}'_p)=\hat z_p({\mathbf {p}}'_o) \pm \hat z_o({\mathbf {p}}).
		\label{eq.DepthGradients}
	\end{equation}
	As the pixels ${\mathbf {p}}'_p$ are regarded as outliers in non-colinear update, we replace $\hat z_o({\mathbf {p}}'_p)$ with $\hat z_o({\mathbf {p}}'_d)$ or $\hat z_o({\mathbf {p}})$ based on the state variable. The depth gradient update function $\bm{\mathcal R}$ in (\ref{def.R}) can be rewritten as follows:
	\begin{equation}
		\begin{split}
			& \hat z_{u}^{\left(k\right)}({\mathbf {p}}),\hat z_{v}^{\left(k\right)}({\mathbf {p}}) = \\
			&    \begin{cases}
				\hat z_p^{\left(k-1\right)}({\mathbf {p}}),      
				&\!\!\!\!\!\!\!\!\!\!\mbox{if }\bm s_u^{\left( k \right)},\bm s_v^{\left( k \right)}  =  \bm 0\\
				\hat z_p^{\left(k-1\right)}({\mathbf {p}})\pm\frac{1}{k+1}\left[z({\mathbf {p}}'_p)-z({\mathbf {p}})\right]\\-\frac{1}{k+1}\hat z_p^{\left(k-1\right)}({\mathbf {p}}'_p),
				&\!\!\!\!\!\!\!\!\!\!\mbox{if }\bm s_u^{\left( k \right)},\bm s_v^{\left( k \right)}  =  {\mathbf {p}}'_p-{\mathbf {p}}\\
				\hat z_p^{\left(k-1\right)}({\mathbf {p}}'_o),  
				&\!\!\!\!\!\!\!\!\!\!\mbox{if }\bm s_u^{\left( k \right)},\bm s_v^{\left( k \right)}  =  {\mathbf {p}}'_o-{\mathbf {p}}\\
				\hat z_p^{\left(k-1\right)}({\mathbf {p}}'_o)\pm\\\left[\hat z_o^{\left(k-1\right)}({\mathbf {p}}) -\hat z_o^{\left(k-1\right)}({\mathbf {p}}'_d)\right], 
				&\!\!\!\!\!\!\!\!\!\!\mbox{if }\bm s_u^{\left( k \right)},\bm s_v^{\left( k \right)} = {\mathbf {p}}'_d-{\mathbf {p}}.
			\end{cases}
		\end{split}
		\label{eq.R}
	\end{equation}
	The representations of $\hat z_p$ and $\hat z_o$ are given in the supplement. The four terms in (\ref{eq.R}), in turn, represent 1) the unchanged depth gradient, 2) parallel update via RPI algorithm, 3) orthogonal update, and 4) diagonal update. The orthogonal and diagonal updates are based on the gradient theorem \cite{williamson2004multivariable}.

	\section{Experiments}
	\subsection{Datasets and Evaluation Metrics}
	In this paper, we followed \cite{3F2N} and conducted comprehensive experiments to qualitatively and quantitatively evaluate the performance of our proposed SDA-SNE. More details on the 3F2N datasets\footnote{\url{sites.google.com/view/3f2n/datasets}} are available in \cite{3F2N}. Moreover, since range sensor data are typically noisy, we add random Gaussian noise of different variances $\sigma$ to the original 3F2N datasets to compare the robustness between our proposed SDA-SNE and other SoTA SNEs.
	
	Two evaluation metrics are used to quantify the accuracy of SNEs:  the average angular error (AAE) \cite{lu2017symps}: 
	\begin{equation}
		e_{\text{A}}(\mathcal{M})=\frac{1}{P}\sum \limits_{k=1}^{P} \phi_k,\,\,\,\,\,\,
		\phi_k=\mathrm{cos^{-1}}\left(\frac{\langle{\mathbf n}_k,\hat{\mathbf n}_k\rangle}{||{\mathbf n}_k||_2||\hat{\mathbf n}_k||_2}\right),
		\label{eq.ea}
	\end{equation}
	and the proportion of good pixels (PGP) \cite{lu2017symps}:
	\begin{equation}
		e_{\text{P}}(\mathcal{M})=\frac{1}{P}\sum \limits_{k=1}^{P}\delta \left(\phi_k,\varphi_k\right),\,\,\,\,\,\,
		\delta=\begin{cases}
			0&(\phi_k>\varphi)\\
			1&(\phi_k\le\varphi)
		\end{cases},
		\label{eq.ep}
	\end{equation}
	where $P$ denotes the total number of pixels used for evaluation, $\varphi$ denotes the angular error tolerance, $\mathcal{M}$ represents the given SNE,  and ${\mathbf n}_k$ and $\hat{\mathbf n}_k$ represent the ground-truth and estimated surface normals, respectively. 
	
	In addition to the above-mentioned evaluation metrics, we also introduce a novel evaluation metric, referred to as cross accuracy ratio (CAR), to depict the performance comparison between two SNEs as follows:
	\vspace{-5pt}
	\begin{equation}
		\vspace{-5pt}
		r_{\text{A}}(\mathcal{M}_1,\mathcal{M}_2)=\frac{e_{\text{A}}(\mathcal{M}_1)}{e_{\text{A}}(\mathcal{M}_2)}.
		\label{eq.ra}
	\end{equation}
	$\mathcal{M}_2$ outperforms $\mathcal{M}_1$ in terms of $e_\text{A}$ if $r_{\text{A}}>1$, and vice versa if $0<r_{\text{A}}<1$. 
	
	\subsection{Implementation Details} 
	\label{sec.imp}
	As discussed in Sec. \ref{sec:Methodology}, initial depth gradients can be estimated by convolving a depth image with image gradient filters, \eg, Sobel \cite{Sobel}, Scharr \cite{Scharr}, Prewitt \cite{Prewitt}, \etc. The second-order filters, \eg, Laplace \cite{Laplace}, can be used to estimate the local depth gradient smoothness $\hat z_{uu}$ and $\hat z_{vv}$. We utilize a finite forward difference (FFD) kernel, $\emph{i.e.}, [0,-1,1]$, as well as a finite backward difference (FBD) kernel, $\emph{i.e.}, [-1,1,0]$ to initialize (coarse) $\hat z_u^{\left( 0 \right)}$ and $\hat z_v^{\left( 0 \right)}$, and use a finite Laplace (FL) kernel, \emph{i.e.}, $[-1,2,-1]$ to estimate $\hat z_{uu}$ and $\hat z_{vv}$. We also conduct experiments w.r.t. different number of iterations to quantify the performance of our proposed SDA-SNE. By implementing the optimum maximum iteration, the trade-off between the speed and accuracy of our algorithm can be significantly minimized. 

	\subsection{Algorithm Convergence and Computational Complexity}
	Our proposed SDA-SNE converges when the state variable and surface normal estimation remains stable. We can obtain the lower bound of the smoothness energy components as follows: 
	\begin{equation}
		E_u({\mathbf {p}}),E_v({\mathbf {p}}) \ge\min\limits_{{\mathbf {p}}'\in \mathcal N_1({\mathbf {p}})}\Big\{|\hat z_{uu}({\mathbf {p}}')|,|\hat z_{vv}({\mathbf {p}}')|\Big\}
		\ge 0.
		\label{eq.C.lb}
	\end{equation}
    The smoothness energy decreases monotonously after each iteration. Therefore, all state variables stop transferring after finite iterations (the smoothness energy converges to a global minimum). 
	
	The computational complexity of our proposed PRI algorithm is $\mathcal{O}(n)$ ($n$ denotes the number of interpolated pixels), when computing the depth gradient using (\ref{eq.R}). Therefore, RPI is much more efficient than Lagrange or Newton polynomial interpolation whose computational complexity is $\mathcal{O}(n^2)$.
	
	Furthermore, when the image resolution is $M\times N$ pixels, the computational complexity of our proposed SDA-SNE is $\mathcal{O}(lMN)$, where $l$ denotes the number of iterations in DP. In most cases, $l\ll M,N$ because DP  automatically stops when $|\hat z_{uu}|$ and $|\hat z_{vv}|$ no longer decrease. If we set a limit on $l$, SDA-SNE's computational complexity can be reduced to $\mathcal{O}(MN)$, which is identical to the computational complexities of 3F2N \cite{3F2N} and CP2TV \cite{CP2TV}.
	
	\subsection{Performance Evaluation}
\begin{figure*}[!t]  
		\centering
		\includegraphics[width=0.960\textwidth]{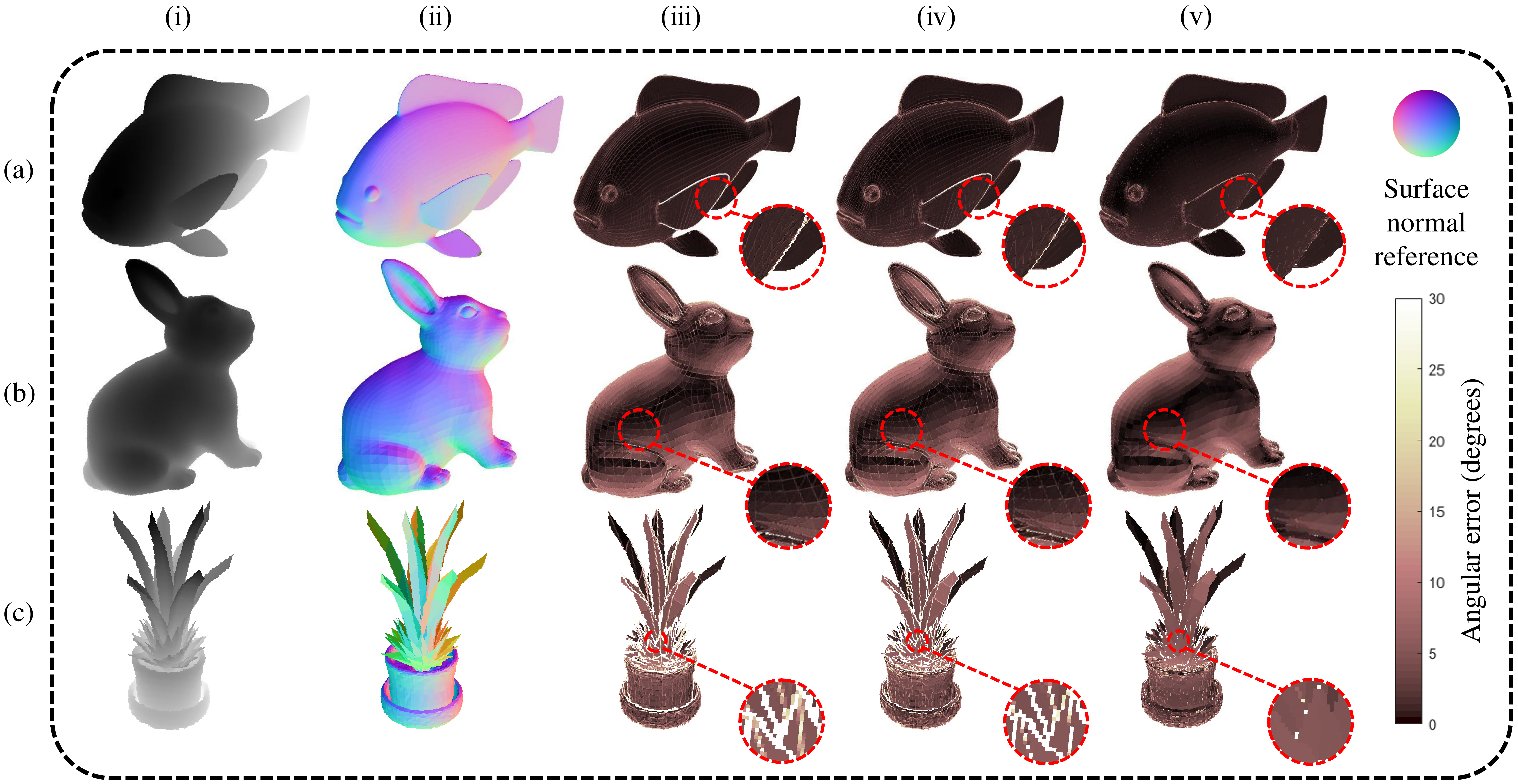}
		\caption{Examples of experimental results: Columns (\romannumeral1)-(\romannumeral5) show the depth images, surface normal ground truth, and the experimental results obtained using 3F2N, CP2TV, and our proposed SDA-SNE, respectively. Rows (a)-(c) show the results of the easy, medium, and hard datasets, respectively.}
		\label{fig.error}
	\end{figure*}

	As our proposed multi-directional DP algorithm mainly aims at improving the performance of depth gradient estimation, it is compatible with any depth-to-normal SNEs. The AAE scores of SoTA depth-to-normal SNEs and such SNEs using our depth gradient estimation strategy are given in Table \ref{table.r}. It can be observed that by using our proposed depth gradient estimation strategy, the AAE scores of such depth-to-normal SNEs decrease by about 20-60\%. Since SDA-SNE based on CP2TV performs better than SDA-SNE based on 3F2N, we use CP2TV to estimate $n_z$ in the following experiments. The qualitative comparison among these SNEs is shown in Fig. \ref{fig.error}. It can be observed that our proposed SDA-SNE significantly outperforms 3F2N and CP2TV near/on discontinuities.
	
	\begin{table}[htb]
		\setlength\tabcolsep{6pt} 
		\begin{center}
			\vspace{0in}
			\footnotesize
			\caption{Comparison of $e_\text{A}$ (degrees) between 3F2N and CP2TV w/ and w/o our depth gradient estimation strategy leveraged.}
			\label{table.r}
			\setlength\arrayrulewidth{0.7pt}
			\begin{tabular}{l|cc|cc}
				\hline
				\multicolumn{1}{c|}{\multirow{2}*{Datasets} }                                                                                   
				& \multicolumn{2}{c|}{3F2N}           & \multicolumn{2}{c}{CP2TV}                                                 \\
				
				\cline{2-5}

				& w/o SDA                                          & w/ SDA                               	& w/o SDA                                      & w/ SDA                            \\

				\hline
				
				Easy 
				& 1.657    & \textbf{0.782}    
				& 1.686    & \textbf{0.677}   
				
				\\
				Medium     
				& 5.686    & \textbf{4.535}   
				& 6.015    & \textbf{4.379}   
				
				\\ 
				Hard
				& 15.315   & \textbf{9.237}  
				& 13.819   & \textbf{8.098}   
				
				\\
				
				\hline
			\end{tabular}
		\end{center}
	\end{table}
	
	\begin{table*}[!t]
		\setlength\tabcolsep{5pt} 
		\begin{center}
			\vspace{0in}
			\footnotesize
			\caption{ Comparison of $e_\text{A}$ and $e_\text{P}$ (with respect to different $\varphi$) among SoTA SNEs on the 3F2N datasets \cite{3F2N}.}
			\label{table.eleven_methods}
			\setlength\arrayrulewidth{0.7pt}
			\begin{tabular}{l|ccc|ccc|ccc|ccc}
				\hline
				\multicolumn{1}{c|}{\multirow{3}*{Method}} & \multicolumn{3}{c|}{$e_\text{A}$ (degrees) $\downarrow$} &  \multicolumn{9}{c}{$e_\text{p}$ $\uparrow$}                                                                                                                                                                                                                                      \\
				\cline{2-13}
				& \multicolumn{1}{c}{\multirow{2}*{ Easy}}              & \multicolumn{1}{c}{\multirow{2}*{ Medium}}    & \multicolumn{1}{c|}{\multirow{2}*{ Hard}}         &  \multicolumn{3}{c|}{\multirow{1}*{Easy}} & \multicolumn{3}{c|}{\multirow{1}*{Medium}} & \multicolumn{3}{c}{\multirow{1}*{Hard}}                                                                                                 \\
				\cline{5-13}
				&
				&                                             &                                              & $\varphi$=10$^\circ$                       & $\varphi$=20$^\circ$                       & $\varphi$=30$^\circ$ & $\varphi$=10$^\circ$ & $\varphi$=20$^\circ$ & $\varphi$=30$^\circ$ 
				& $\varphi$=10$^\circ$                       & $\varphi$=20$^\circ$                       & $\varphi$=30$^\circ$\\
				\hline

				PlaneSVD \cite{klasing2009realtime}  
				& 2.07
				& 6.07
				& 17.59
				& 0.9648                                       
				& 0.9792                      
				& 0.9855                   
				& 0.8621                                      
				& 0.9531                                      
				& 0.9718                                      
				& 0.6202                                      
				& 0.7394                                      
				& 0.7914

				\\
				
				PlanePCA \cite{jordan2014quantitative}   
				& 2.07
				& 6.07
				& 17.59
				& 0.9648
				& 0.9792         
				& 0.9855             
				& 0.8621                             
				& 0.9531 
				& 0.9718     
				& 0.6202             
				& 0.7394         
				& 0.7914

				\\

				VectorSVD \cite{klasing2009comparison}   
				& 2.13
				& 6.27
				& 18.01
				& 0.9643                                       
				& 0.9777                      
				& 0.9846                   
				& 0.8601                                
				& 0.9495                
				& 0.9683                 
				& 0.6187           
				& 0.7346  
				& 0.7848

				\\

				AreaWeighted \cite{klasing2009comparison} 
				& 2.20
				& 6.27
				& 17.03
				& 0.9636                                       
				& 0.9753                      
				& 0.9819                   
				& 0.8634               
				& 0.9504                      
				& 0.9665                     
				& 0.6248                     
				& 0.7448                      
				& 0.7977
				
				\\
				
				AngleWeighted \cite{klasing2009comparison}  
				& 1.79
				& 5.67
				& 13.26
				& 0.9762                            
				& 0.9862          
				& 0.9893         
				& 0.8814               
				& 0.9711               
				& 0.9809            
				& 0.6625                        
				& 0.8075            
				& 0.8651
				
				\\
				FALS \cite{badino2011fast}
				& 2.26                                     
				& 6.14                      
				& 17.34  
				& 0.9654                                       
				& 0.9794                      
				& 0.9857                   
				& 0.8621     
				& 0.9547        
				& 0.9731         
				& 0.6209           
				& 0.7433            
				& 0.7961

				\\

				SRI \cite{badino2011fast}   
				& 2.64                                     
				& 6.71                      
				& 19.61  
				& 0.9499                                       
				& 0.9713                      
				& 0.9798                   
				& 0.8431          
				& 0.9403   
				& 0.9633         
				& 0.5594  
				& 0.6932 
				& 0.7605
				
				\\

				LINE-MOD \cite{hinterstoisser2011gradient}  
				& 2.64                                     
				& 6.71                      
				& 19.61
				& 0.8542                                      
				& 0.9085                                       
				& 0.9343         
				& 0.7277   
				& 0.8803 
				& 0.9282   
				& 0.3375 
				& 0.4757 
				& 0.5636

				\\
				
				SNE-RoadSeg \cite{fan2020sne}   
				& 2.04                                     
				& 6.28                      
				& 16.37
				& 0.9693                                       
				& 0.9810                      
				& 0.9871                   
				& 0.8618               
				& 0.9512               
				& 0.9725               
				& 0.6226               
				& 0.7589               
				& 0.8113

				\\

				3F2N \cite{3F2N}  
				& 1.66                                     
				& 5.69                      
				& 15.31
				& 0.9723                                       
				& 0.9829                      
				& 0.9889                          
				& 0.8722      
				& 0.9600      
				& 0.9766      
				& 0.6631
				& 0.7821 
				& 0.8289
				
				\\
				CP2TV \cite{CP2TV}     
				
				& 1.69                                     
				& 6.02                      
				& 13.82
				& 0.9740                                       
				& 0.9843                      
				& 0.9899                          
				& 0.8512     
				& 0.9554      
				& 0.9755      
				& 0.6840
				& 0.8099 
				& 0.8562
				\\
				\hline
				\rowcolor{gray!20}
				SDA-SNE (iteration = $\infty$)
				& 0.68
				& \textbf{4.38}
				& \textbf{8.10}
				& \textbf{0.9947}
				& 0.9982
				& 0.9991
				& \textbf{0.9075}
				& \textbf{0.9868}
				& \textbf{0.9939}
				& \textbf{0.8035}
				& \textbf{0.9254}
				& \textbf{0.9461}
				\\
				
				\rowcolor{gray!20}
				SDA-SNE (iteration = $3$)
				& \textbf{0.67}
				& \textbf{4.38}
				& 8.14
				& \textbf{0.9947}
				& \textbf{0.9983}
				& \textbf{0.9992}
				& \textbf{0.9075}
				& 0.9867
				& 0.9938
				& 0.8027
				& 0.9174
				& 0.9453
				\\
				\hline
			\end{tabular}
		\end{center}
	\end{table*}
	
	\begin{figure}[!t]  
		\centering
		\includegraphics[width=0.46\textwidth]{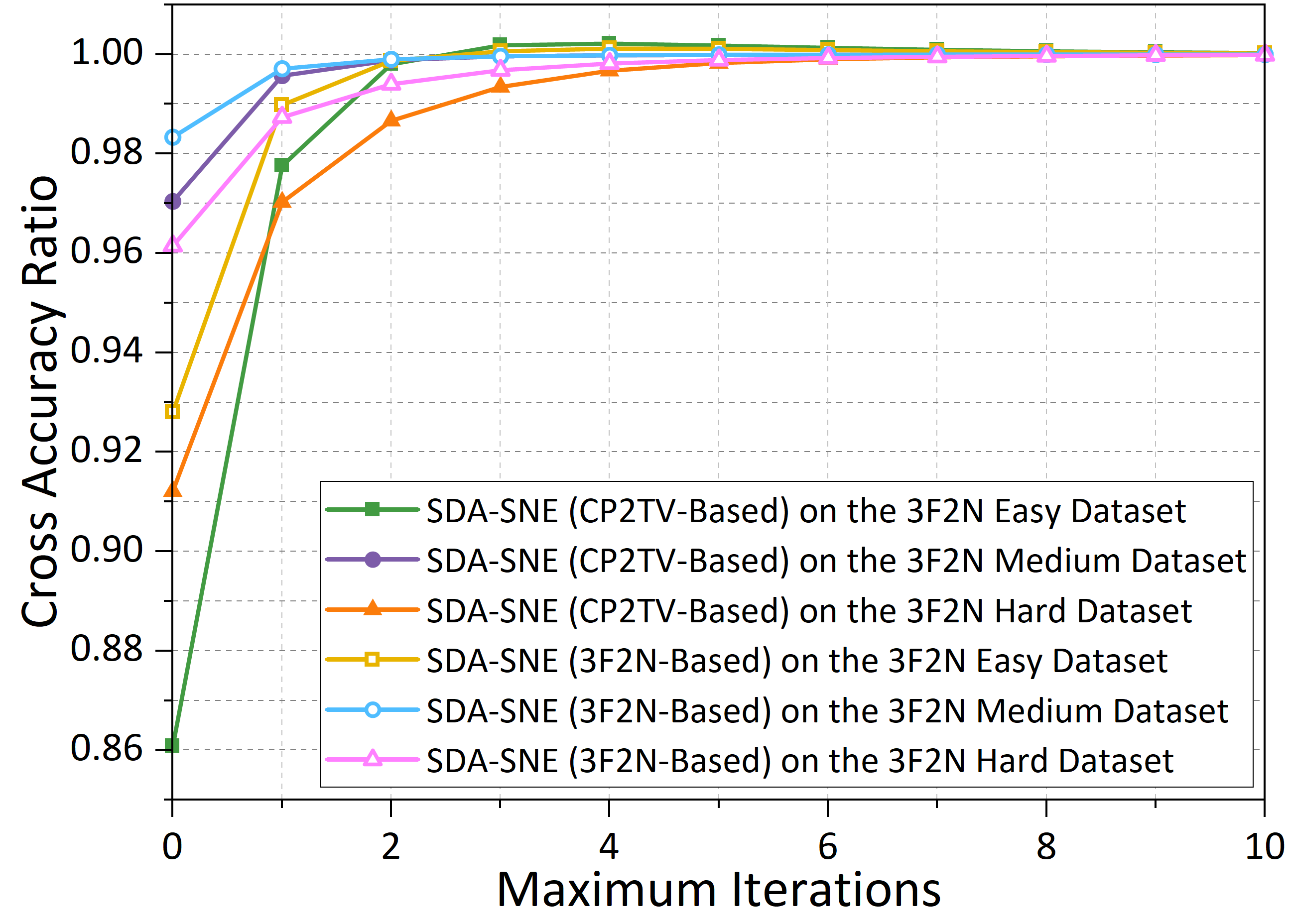}
		\caption{CAR comparison between CP2TV-based and 3F2N-based SDA-SNEs w.r.t. different maximum iterations.}
		\label{fig.base}
	\end{figure}
	
	Additionally, we compare the performance of our proposed SDA-SNE w.r.t. a collection of maximum iterations. Fig. \ref{fig.base} shows the CAR scores achieved by CP2TV-based and 3F2N-based SDA-SNEs on the 3F2N easy, medium, and hard datasets. It can be observed that the performance of SDA-SNE saturates after only several iterations. The accuracy increases by less than 5\% when the maximum iteration is set to infinity. Therefore, the maximum iteration of multi-directional DP can be set to 3 to minimize its trade-off between speed and accuracy. Moreover, it can be observed that our proposed SDA-SNE converges (the CAR score reaches the maximum) with more iterations on the hard dataset than on the easy and medium datasets. This is probably due to the fact that the hard dataset possesses more discontinuities than the easy and medium datasets. As a result, more distant pixels are required to be introduced to yield better depth gradients.

	\begin{figure}[!t] 
		\centering
		\includegraphics[width=0.46\textwidth]{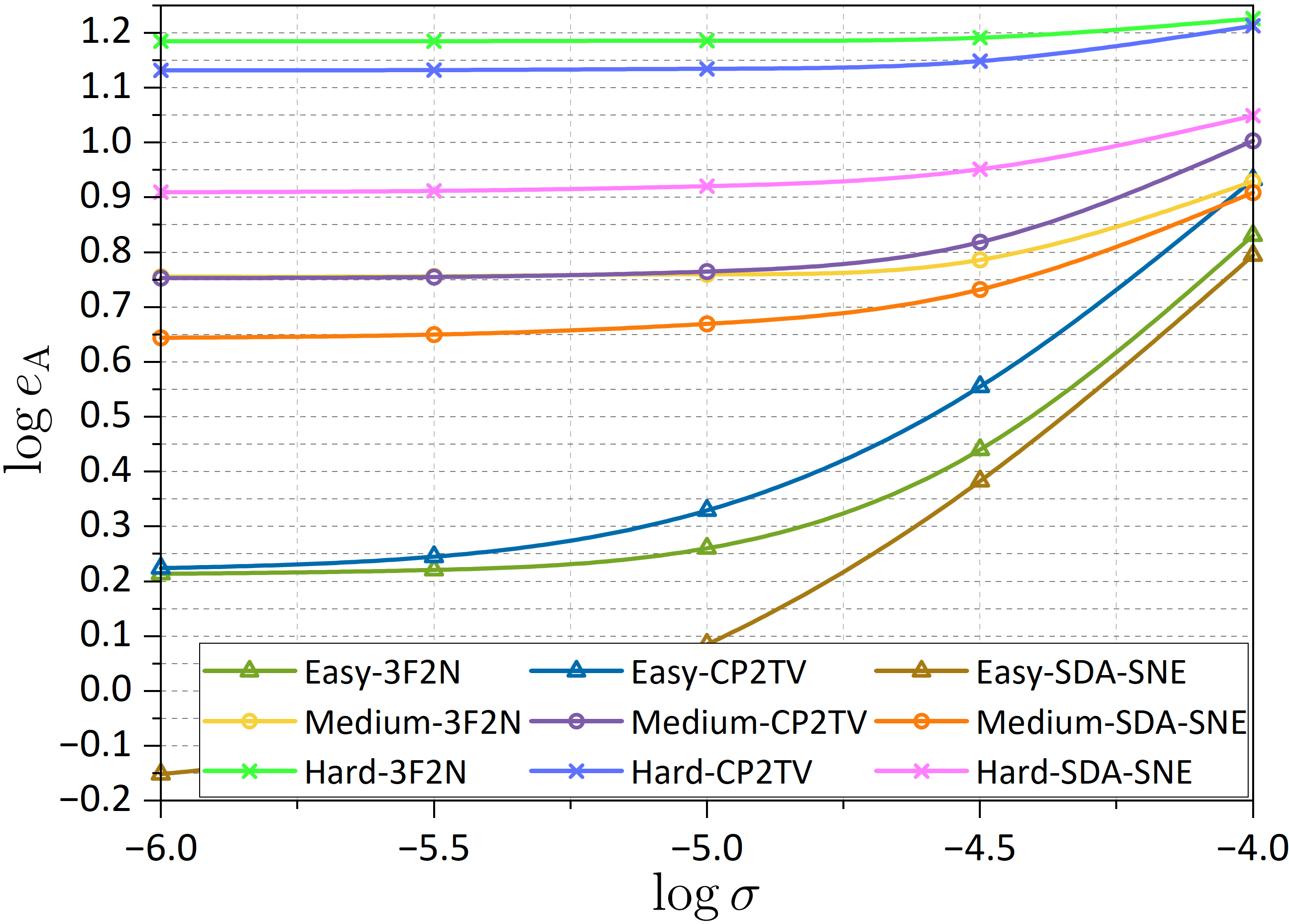}
		\caption{AAE comparison among 3F2N, CP2TV, and SDA-SNE on the 3F2N datasets with different levels of Gaussian noise added.}
		\label{fig.noise}
	\end{figure}
	
	Moreover, we compare our proposed SDA-SNE with all other SoTA SNEs presented in Sec. \ref{sec:RelatedWork}.  $e_\text{A}$ and $e_\text{P}$ of all SNEs on the 3F2N easy, medium, and hard datasets are given in Table \ref{table.eleven_methods}. The ${e_\text{A}}$ scores achieved by SDA-SNE are less than $1^\circ$ (easy), $5^\circ$ (medium), and $9^\circ$ (hard), respectively. The ${e_\text{P}}$ scores (tolerance: $10^\circ$) achieved by SDA-SNE are about 100\% (easy), 90\% (medium), and 80\% (hard), respectively. These results suggest that our proposed SDA-SNE performs significantly better than all other SoTA SNEs, no matter whether an iteration limit is added or not. 
	
	Furthermore, we compare the robustness (to random Gaussian noise) of our proposed SDA-SNE with two SoTA depth-to-normal SNEs (3F2N \cite{3F2N} and CP2TV \cite{CP2TV}), as shown in Fig. \ref{fig.noise}. The logarithms of AAE scores and Gaussian variances are used to include the results of different datasets into a single figure. It is evident that (1) the $e_\text{A}$ scores achieved by all SNEs increase monotonically with the increasing noise level, and (2) our proposed SDA-SNE outperforms 3F2N and CP2TV at different noise levels. In addition, the compared SNEs' performance on the easy dataset degrades more dramatically than the medium and hard datasets, as the original easy dataset contains much fewer discontinuities. Although the performance of these depth-to-normal SNEs becomes remarkably similar, with the increase in noise level, our proposed SDA-SNE consistently outperforms 3F2N and CP2TV. This further demonstrates the superior robustness of our algorithm over others. 

	\section{Conclusion}
	This paper presented a highly accurate discontinuity-aware surface normal estimator, referred to as SDA-SNE. Our approach computes surface normals from a depth image by iteratively introducing adjacent co-planar pixels using a novel multi-directional dynamic programming algorithm. To refine the depth gradient in each iteration, we introduced a novel recursive polynomial interpolation algorithm with high computational efficiency. Our proposed depth gradient estimation approach is compatible with any depth-to-normal surface normal estimator, such as 3F2N and CP2TV. To evaluate the accuracy of our proposed surface normal estimator, we conducted extensive experiments on both clean and noisy datasets. Our proposed SDA-SNE achieves the highest accuracy on clean 3F2N datasets ($0.68^\circ$, $4.38^\circ$, $8.10^\circ$ on the easy, medium, and hard datasets, respectively), outperforming all other SoTA surface normal estimators. It also demonstrates high robustness to different levels of random Gaussian noise. Additional experimental results suggest that our proposed SDA-SNE can achieve a similar performance when reducing the maximum iteration of multi-directional dynamic programming to $3$. This ensures the high efficiency of our proposed SDA-SNE in various computer vision and robotics applications requiring real-time performance.

	\section{Acknowledgements}
	
	This work was supported by the National Key R\&D Program of China, under grant No. 2020AAA0108100, awarded to Prof. Qijun Chen. This work was also supported by the Fundamental Research Funds for the Central Universities, under projects No. 22120220184, No. 22120220214, and No. 2022-5-YB-08, awarded to Prof. Rui Fan. 
	
	\clearpage
	\balance
	{\small
		\bibliographystyle{ieee_fullname}

	}
	\clearpage
	
	{\Large \textbf{\center{-- Supplementary Material --\\}}}
	
	\section{Details on Path Discontinuity Norm}
	Plugging the Taylor expansions of $z_u$ and $z_v$ into (\ref{eq.PathIntegral.1}) and (\ref{eq.PathIntegral.2}) results in the upper bound of the depth gradient estimation error w.r.t. collinear transfer and non-collinear transfer. The depth gradient estimation error can be associated with our introduced PD norm (detailed in Sec. \ref{sec:co} and \ref{sec:nonco}). Furthermore, we compare PD norm with total variation (TV) norm \cite{CISP2010}, as detailed in Sec. \ref{sec:tv}.
	\subsection{Depth Gradient Error w.r.t. Collinear Transfer}
	\label{sec:co}
	In collinear transfer, (\ref{eq.PathIntegral.1}) can be rewritten as follows:
	\begin{equation}
		\begin{split}
			& z({\mathbf {p}}')-z(\mathbf {p})=\\
			& (u_1-u_0)z_u(\mathbf {p})+\int_{u_0}^{u_1} (u-u_0)z_{uu}\,du,\mbox{ or }\\
			& z({\mathbf {p}}')-z(\mathbf {p})=\\
			& (v_1-v_0)z_v(\mathbf {p})+\int_{v_0}^{v_1} (v-v_0)z_{vv}\,dv.
		\end{split}
		\label{supp.Co.1}
	\end{equation}
	Omitting the second-order terms results in the following collinear transfer error expressions:
	\begin{equation}
		\begin{split}
			& \varepsilon_{\text{C}}=\left\lvert \int_{u_0}^{u_1} (u-u_0)z_{uu}\,du\right\rvert,\mbox{ or }\\
			& \varepsilon_{\text{C}}=\left\lvert \int_{v_0}^{v_1} (v-v_0)z_{vv}\,dv\right\rvert.
		\end{split}
		\label{supp.Co.2}
	\end{equation}
	As $|u_0-u_1| \le 1$ and $|v_0-v_1| \le 1$, PD norm is the upper bound of the collinear transfer error $\varepsilon_{\text{C}}$:
	\begin{equation}
		\varepsilon_{\text{C}}\le ||z||_{\text{PD}},\,\,\,\,\,\,\,\,\mathcal L : \mathbf {p} \to {\mathbf {p}}'.
		\label{supp.Co.3}
	\end{equation}
	
	\subsection{Depth Gradient Error w.r.t. Non-Collinear Transfer}
	\label{sec:nonco}
	In non-collinear transfer, (\ref{eq.PathIntegral.2}) can be revised as follows:
	\begin{equation}
		\begin{split}
			& (u_1-u_0)z_u(\mathbf {p})+(v_1-v_0)z_v({\mathbf {p}}')\\
			& +\int_{\mathcal L_1}(u-u_0)z_{uu}\,du+(v-v_1)z_{vv}\,dv\\
			=\,\,
			& (v_1-v_0)z_v(\mathbf {p})+(u_1-u_0)z_u({\mathbf {p}}')\\
			& +\int_{\mathcal L_2}(u-u_1)z_{uu}\,du+(v-v_0)z_{vv}\,dv.
		\end{split}
		\label{supp.NonCo.1}
	\end{equation}
	PD norm is also the upper bound of the non-collinear transferring error $\varepsilon_{\text{N}}$ w.r.t. different DP paths:
	\begin{equation}
		\begin{split}
			& \varepsilon_{\text{N}} \\
			& = \left\lvert \int_{\mathcal L_1}(u-u_0)z_{uu}\,du+(v-v_1)z_{vv}\,dv-\right.\\
			& \left.\int_{\mathcal L_2}(u-u_1)z_{uu}\,du+(v-v_0)z_{vv}\,dv \right\rvert\\ 
			& \le \int_{\mathcal L_1}| z_{uu}\,du|+|z_{vv}\,dv|+\int_{\mathcal L_2}|z_{uu}\,du|+|z_{vv}\,dv|\\
			& =\left\lvert\left\lvert z \right\rvert\right\rvert_{\text{PD}},\,\,\,\,\,\,\,\,\mathcal L =\mathcal L_1 + \mathcal L_2: \mathbf {p} \to {\mathbf {p}}'\to \mathbf {p}.
		\end{split}
		\label{supp.NonCo.2}
	\end{equation}
	
	Therefore, the PD norm determines the upper bound of the depth gradient estimation error. Optimizing the PD norm in (\ref{eq.PDN.C}) is equivalent to minimizing the error of $z_u$ and $z_v$ in (\ref{supp.Co.2}) and (\ref{supp.NonCo.2}).
	
	\subsection{Comparison between PD Norm and TV Norm}
	\label{sec:tv}
	TV norm \cite{CISP2010} is widely used to measure depth discontinuity. Compared to TV norm, our proposed PD norm introduces the second-order partial derivative of depth, demonstrating better awareness of ridges. We conduct additional experiments on the 3F2N datasets to compare the accuracy of our method using PD and TV norms, respectively. As shown in Table \ref{table.supp.1}, the accuracy achieved when using PD norm is higher than that achieved when using TV norm.
	
	\begin{table}[htb]
		\begin{center}
			\footnotesize
			\caption{Comparison of $e_\text{A}$ when using TV and PD norms.}
			\label{table.supp.1}
			\setlength\tabcolsep{7pt} 
			\setlength\arrayrulewidth{0.8pt}
			\begin{tabular}{l|c|c|c}
				\hline
				\multicolumn{1}{c|}{\multirow{2}*{Method} }                                                                                   
				& \multicolumn{3}{c}{$e_\text{A}$ (degrees) $\downarrow$}                                                    \\
				
				\cline{2-4}

				& Easy                                         & Medium                              	& Hard                                                   \\

				\hline
				
				SDA-SNE (with TV norm)
				& 1.37  
				& 5.14   
				& 10.05 
				
				\\ 
				SDA-SNE (with PD norm)
				& \textbf{0.67}    
				& \textbf{4.38}   
				& \textbf{8.14}
				
				\\
				
				\hline
			\end{tabular}
		\end{center}
	\vspace{-2em}
	\end{table}
	
		\begin{figure*}[t!]  
		\centering
		\includegraphics[width=0.995\textwidth]{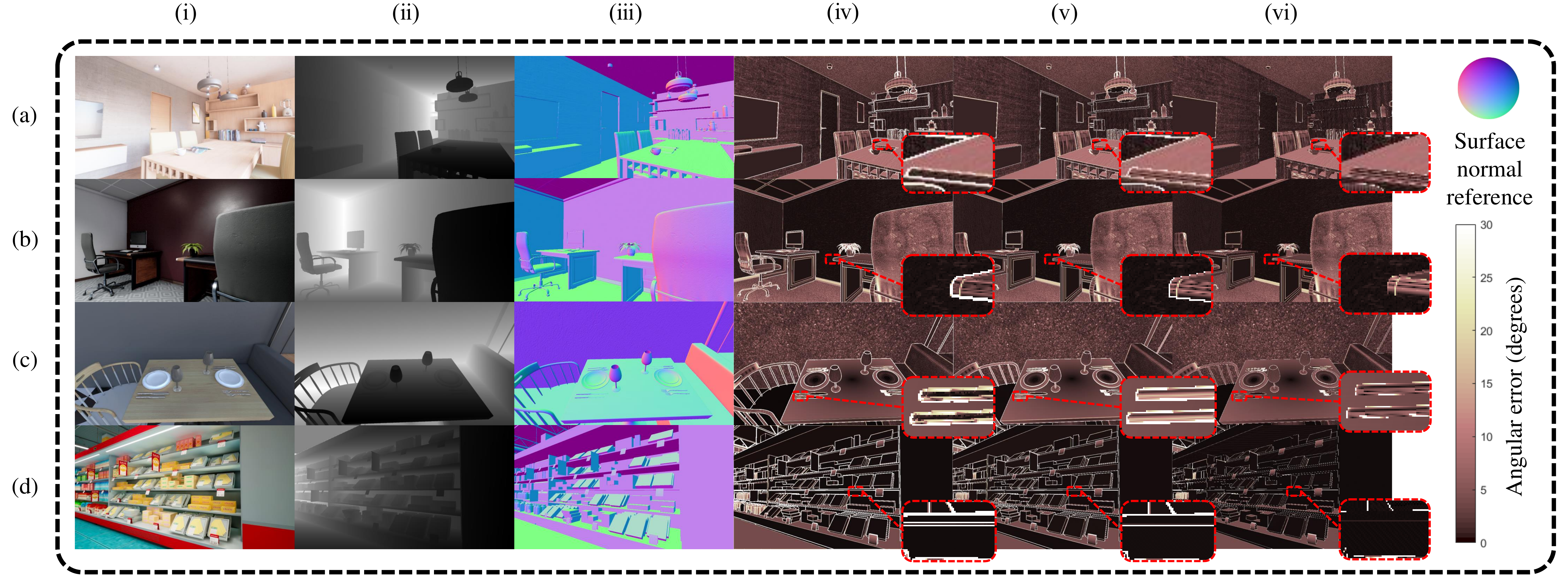}
		\caption{Examples of experimental results: Columns (\romannumeral1)-(\romannumeral6) show the RGB images, depth images, surface normal ground truth, and the experimental results obtained using 3F2N, CP2TV, and our proposed SDA-SNE, respectively. Rows (a)-(d) show the home, office, restaurant, and store scenarios, respectively.}
		\label{fig.supp}
	\end{figure*}
 
	\section{RPI Theorem}
	\paragraph{Theorem:} 
	If the $n$-th order polynomial $f(u)^{\left( n \right)}\in P^{n}[u]$ is interpolated by $\{(u_0,z_0),(u_1,z_1),\cdots,(u_{n},z_{n})\}$, it can be expanded into a recursive form as follows: 
	\begin{equation}
		f^{\left( n \right)}=\frac{u-u_n}{u_0-u_n}f_0^{\left( n-1 \right)}+\frac{u-u_0}{u_n-u_0}f_1^{\left( n-1 \right)}, 
		\label{eq.lemma}
	\end{equation}
	where $f_0^{\left( n-1 \right)}$ and $f_1^{\left( n-1 \right)}$ are respectively interpolated by the first and the last $n$ points.

	\paragraph{Proof}
	\hspace*{\fill} \\
	(a) Substituting  $u=u_0$ into (\ref{eq.lemma}), we can obtain:
	\begin{equation}
	    \begin{split}
		\frac{u-u_n}{u_0-u_n} &=1,\frac{u-u_0}{u_n-u_0}  =0,\\
		f^{\left( n \right)}(u_0) &=f_0^{\left( n-1 \right)}(u_0)=z_0.
		\end{split}
	\end{equation}
	(b) Substituting  $u=u_n$ into (\ref{eq.lemma}), we can obtain:
	\begin{equation}
	    \begin{split}
		\frac{u-u_n}{u_0-u_n} &=0,\frac{u-u_0}{u_n-u_0}  =1,\\
		f^{\left( n \right)}(u_n) &=f_1^{\left( n-1 \right)}(u_n)=z_n.
		\end{split}
	\end{equation}
	(c) Substituting  $u=u_1,\cdots, u_{n-1}$ into (\ref{eq.lemma}), we can obtain:
	\begin{equation}
		\begin{split}
		\frac{u-u_0}{u_n-u_0} &+\frac{u-u_n}{u_0-u_n}=1,\\
		f^{\left( n \right)}(u_i) &=f_0^{\left( n-1\right)}(u_i)=f_1^{\left( n-1 \right)}(u_i)=z_i, 
		\end{split}
	\end{equation}
	where $i$ is an integer between $1$ and $n-1$.
	
		\section{Additional Experiments on Noisy Data}
	We also conduct additional experiments on the IRS dataset \footnote{\url{github.com/HKBU-HPML/IRS}} \cite{IRS} to further validate our proposed SDA-SNE's robustness to noise. The IRS dataset contains four indoor scenarios (home, office, restaurant, and store) \cite{IRS}. The comparison among 3F2N \cite{3F2N}, CP2TV \cite{CP2TV}, and our proposed SDA-SNE in terms of $e_\text{A}$ is given in Table. \ref{table.supp.2}. It can be observed that our proposed SDA-SNE also outperforms 3F2N and CP2TV on the IRS dataset ($e_\text{A}$ is lower by about 40\% and 30\%, respectively). These results suggest that SDA-SNE performs robustly on real noisy data under different illumination conditions. Qualitative comparison among 3F2N, CP2TV, and SDA-SNE is given in Fig. \ref{fig.supp}. It can be observed that compared to 3F2N and CP2TV, SDA-SNE performs much better near/on discontinuities.
	\begin{table}[htb]
		\begin{center}
			\footnotesize
			\caption{Comparison of $e_\text{A}$ (degrees) among 3F2N, CP2TV, and our proposed SDA-SDA on the IRS dataset.}
			\label{table.supp.2}
			\setlength\arrayrulewidth{0.7pt}
			\begin{tabular}{l|c|c|c|c}
				\hline
				\multicolumn{1}{c|}{\multirow{2}*{Method} }                                                                                   
				& \multicolumn{4}{c}{$e_\text{A}$ (degrees) $\downarrow$}                                                    \\
				
				\cline{2-5}

				& Home                                         & Office                              	& Restaurant                                      & Store                            \\

				\hline
				
				3F2N \cite{3F2N}
				& 6.2665    & 6.9919    
				& 8.2611    & 7.4258  
				
				\\
				CP2TV \cite{CP2TV}
				& 5.6835    & 6.2551
				& 7.3373    & 6.2356  
				
				\\ 
				\hline
				\rowcolor{gray!20}
				SDA-SDA (ours)
				& \textbf{3.9073}   & \textbf{4.4796}  
				& \textbf{5.2782}   & \textbf{4.4245}
				
				\\
				
				\hline
			\end{tabular}
		\end{center}
	\end{table}

		\section{Detailed Representations of Subscripts $p$, $o$, and $d$ in Multi-Directional DP}
	The representations of $E_p^{\left( k-1 \right)}$, $E_o^{\left( k-1 \right)}$, $\hat z_{pp}$, and $\hat z_{oo}$ in (\ref{eq.C}) are shown as follows:
	\begin{equation}
		\begin{split}
			E_p^{\left( k-1 \right)}=
			\begin{cases}
				E_u^{\left( k-1 \right)}, 
				& \mbox{when computing $E_u^{\left( k \right)}$}\\
				E_v^{\left( k-1 \right)},
				& \mbox{when computing $E_v^{\left( k \right)}$}
			\end{cases},\\
			E_o^{\left( k-1 \right)}=
			\begin{cases}
				E_v^{\left( k-1 \right)}, 
				& \mbox{when computing $E_u^{\left( k \right)}$}\\
				E_u^{\left( k-1 \right)},
				& \mbox{when computing $E_v^{\left( k \right)}$}
			\end{cases},
			\label{eq.1}
		\end{split}
	\end{equation}
	and
	\begin{equation}
		\begin{split}
			\hat z_{pp}=
			\begin{cases}
				\hat z_{uu}, 
				& \mbox{when computing $E_u^{\left( k \right)}$}\\
				\hat z_{vv},
				& \mbox{when computing $E_v^{\left( k \right)}$}
			\end{cases},\\
			\hat z_{oo}=
			\begin{cases}
				\hat z_{vv}, 
				& \mbox{when computing $E_u^{\left( k \right)}$}\\
				\hat z_{uu},
				& \mbox{when computing $E_v^{\left( k \right)}$}
			\end{cases}.
			\label{eq.2}
		\end{split}
	\end{equation}
	The representations of $\hat z_{p}^{\left( k-1 \right)}$ and $\hat z_{o}^{\left( k-1 \right)}$ in (\ref{eq.R}) are shown as follows:
	\begin{equation}
		\begin{split}
			\hat z_p^{\left( k-1 \right)}=
			\begin{cases}
				\hat z_u^{\left( k-1 \right)}, 
				& \mbox{when computing $\hat z_u^{\left( k \right)}$}\\
				\hat z_v^{\left( k-1 \right)},
				& \mbox{when computing $\hat z_v^{\left( k \right)}$}
			\end{cases},\\
			\hat z_o^{\left( k-1 \right)}=
			\begin{cases}
				\hat z_v^{\left( k-1 \right)}, 
				& \mbox{when computing $\hat z_u^{\left( k \right)}$}\\
				\hat z_u^{\left( k-1 \right)},
				& \mbox{when computing $\hat z_v^{\left( k \right)}$}
			\end{cases}.
			\label{eq.3}
		\end{split}
	\end{equation}
	Given a pixel $\mathbf {p} = [u_0,v_0]^{\top}$, its adjacent pixels $\mathbf {p}'_p$, $\mathbf {p}'_o$, and $\mathbf {p}'_d$ are denoted as follows:
	\begin{equation}
		\begin{split}
			\mathbf {p}'_p &=
			\begin{cases}
				\big[u_0\pm 1, v_0\big]^{\top}, 
				& \mbox{when computing $E_u^{\left( k \right)}$ or $\hat z_u^{\left( k \right)}$}\vspace{+0.3em}\\
				\big[u_0, v_0\pm 1\big]^{\top},
				& \mbox{when computing $E_v^{\left( k \right)}$ or $\hat z_v^{\left( k \right)}$}
			\end{cases},\\
			\mathbf {p}'_o &=
			\begin{cases}
				\big[u_0, v_0\pm 1\big]^{\top},
				& \mbox{when computing $E_u^{\left( k \right)}$ or $\hat z_u^{\left( k \right)}$}\vspace{+0.3em}\\
				\big[u_0\pm 1, v_0\big]^{\top}, 
				& \mbox{when computing $E_v^{\left( k \right)}$ or $\hat z_v^{\left( k \right)}$}
			\end{cases},\\
			\mathbf {p}'_d &=\big[u_0\pm 1, v_0\pm 1\big]^{\top}.
			\label{eq.4}
		\end{split}
	\end{equation}


	\newpage
	\section{Nomenclature}
	\begin{table*}[b!]
		\begin{center}
			\footnotesize
			\renewcommand{\arraystretch}{1.2}
			\begin{tabular}{cll}
				\hline
				
				Notation    
				&     
				& Description 
				\\
				\hline
				
				$\mathbf{p}^C = [x,y,z]^\top$
				& $\in \mathbb{R}^3$  
				& Coordinate of a 3D point in the camera coordinate system
				\\ 
				
				$\mathbf{p} = [u,v]^\top$
				& $\in \mathbb{R}^2$ 
				& Coordinate of a 2D pixel in the pixel coordinate system
				\\
				
				${\mathbf {p}}'$
				& $\in \mathbb{R}^2$ 
				& Adjacent pixel of a given pixel $\mathbf {p}$
				\\
				
				$\mathbf{K}$
				& 
				& Camera intrinsic matrix
				\\
				
				$\nabla z$
				& $\in \mathbb{R}^2$ 
				& Theoretical depth gradient
				\\
				
				$\nabla \hat z$
				& $\in \mathbb{R}^2$ 
				& Estimated depth gradient
				\\
				
				$\mathbf n$
				& $\in \mathbb{R}^3$
				& Theoretical surface normal
				\\
				
				$\hat {\mathbf n}$
				& $\in \mathbb{R}^3$
				& Estimated surface normal
				\\
				
				$\cdot^{\left( k \right)}$
				& 
				& The $k$-th iteration
				\\
				
				$\cdot^{*}$
				& 
				& The optimal value
				\\
				
				$E_u,E_v$
				& $\in \mathbb{R}_+$
				& Smoothness energy components on the $u$-axis and $v$-axis
				\\
				
				$\bm s_u, \bm s_v$
				&  $\in \mathbb{S}:\big\{[i,j]^{\top}\mid i,j\in \{-1,0,1\}\big\}$
				& State variable components on the $u$-axis and $v$-axis
				\\
				
				$\bm{\mathcal T}$
				& $:\mathbb{R}^2_+ \times \mathbb{S}^2 \mapsto \mathbb{R}^2_+$
				& Smoothness energy transfer function
				\\
				
				$\bm{\mathcal R}$
				& $:\mathbb{R}^2 \times \mathbb{S}^2 \mapsto \mathbb{R}^2$
				& Depth gradient update function
				\\
				
				$\Delta_f z$
				& $\in \mathbb{R}^2$ 
				& Forward finite difference of depth on the $u$-axis and $v$-axis
				\\
				
				$\Delta_b z$
				& $\in \mathbb{R}^2$ 
				& Backward finite difference of depth on the $u$-axis and $v$-axis
				\\
				$z_u,z_v$
				& $\in \mathbb{R}$
				& Theoretical first-order partial derivatives of depth on the $u$-axis and $v$-axis
				\\
				
				$\hat z_u,\hat z_v$
				& $\in \mathbb{R}$
				& Estimated first-order partial derivatives of depth on the $u$-axis and $v$-axis
				\\
				
				$z_{uu},z_{vv}$
				& $\in \mathbb{R}$
				& Theoretical second-order partial derivatives of depth on the $u$-axis and $v$-axis
				\\
				
				$\hat z_{uu},\hat z_{vv}$
				& $\in \mathbb{R}$
				& Estimated second-order partial derivatives of depth on the $u$-axis and $v$-axis
				\\
				
				$\cdot_p,\cdot_d,\cdot_o$
				&
				& Parallel, orthogonal, or diagonal to a given axis, respectively
				\\
				
				$||\cdot||_\text{PD}$
				&
				& Path discontinuity norm
				\\
				
				$f^{\left( n \right)}$
				&
				& The $n$-th order polynomial 
				\\
				
				$e_\text{A}$
				& $\in \mathbb{R}_+$
				& Average angular error
				\\
				
				$e_\text{P}$
				&
				& Proportion of good pixels
				\\
				
				$r_\text{A}$
				&
				& Cross accuracy ratio
				\\
				
				$\varepsilon_{\text{C}}$
				& $\in \mathbb{R}_+$ 
				& Collinear transfer error of depth gradient estimation
				\\
				
				$\varepsilon_{\text{N}}$
				& $\in \mathbb{R}_+$ 
				& Non-collinear transfer error of depth gradient estimation
				\\
				
				\hline
			\end{tabular}
			\caption{Nomenclature}
			\label{table.notation}
		\end{center}
	\end{table*}
	
\end{document}